\documentclass[runningheads]{llncs}
\usepackage[english]{babel}
\usepackage{cite}
\usepackage{float}
\usepackage{booktabs} 
\usepackage{graphicx} 
\usepackage[letterpaper,top=2cm,bottom=2cm,left=3cm,right=3cm,marginparwidth=1.75cm]{geometry}
\usepackage{amsmath}
\usepackage{graphicx}
\usepackage[colorlinks=true, allcolors=blue]{hyperref}
\usepackage[font=small]{caption}
\usepackage[font=footnotesize]{caption}

\title{HyboWaveNet: Hyperbolic Graph Neural Networks with Multi-Scale Wavelet Transform for Protein-Protein Interaction Prediction}
\usepackage{fancyhdr}
\author{Qingzhi Yu \and shuai Yan \and Wenfeng Dai\and Xiang Cheng\thanks{Correspondence: School of Information Engineering, Jingdezhen Ceramics University,mail:jx\_chx@126.com} }
\institute{Jingdezhen Ceramics University,Jiangxi,China}
\begin{document}
\maketitle

\begin{abstract}
Protein-protein interactions (PPIs) are fundamental for deciphering cellular functions, disease pathways, and drug discovery. Although existing neural networks and machine learning methods have achieved high accuracy in PPI prediction, their black-box nature leads to a lack of causal interpretation of the prediction results and difficulty in capturing hierarchical geometries and multi-scale dynamic interaction patterns among proteins. To address these challenges, we propose HyboWaveNet, a novel deep learning framework that collaborates with hyperbolic graphical neural networks (HGNNs) and multiscale graphical wavelet transform for robust PPI prediction. Mapping protein features to Lorentz space simulates hierarchical topological relationships among biomolecules via a hyperbolic distance metric, enabling node feature representations that better fit biological a priori.HyboWaveNet inherently simulates hierarchical and scale-free biological relationships, while the integration of wavelet transforms enables adaptive extraction of local and global interaction features across different resolutions. Our framework generates node feature representations via a graph neural network under the Lorenz model and generates pairs of positive samples under multiple different views for comparative learning, followed by further feature extraction via multi-scale graph wavelet transforms to predict potential PPIs. Experiments on public datasets show that HyboWaveNet improves over both existing state-of-the-art methods. We also demonstrate through ablation experimental studies that the multi-scale graph wavelet transform module improves the predictive performance and generalization ability of HyboWaveNet. This work links geometric deep learning and signal processing to advance PPI prediction, providing a principled approach for analyzing complex biological systems The source code is available at https://github.com/chromaprim/HybowaveNet.

\end{abstract}

\section{Introduction}

Protein is a central component of cellular function and biological processes. It is essential for regulating many biological activities in cells \cite{alberts1998cell}. And protein-protein interaction (PPI) is a central issue in biological research, which is essential for understanding cellular functions, disease pathways, and drug discovery. The parsing of PPI networks helps to reveal complex regulatory mechanisms and signaling pathways in biological systems\cite{RN46,RN45,RN44,vidal2011interactome}. Most deep learning-based PPIS prediction methods use mainstream core networks, such as convolutional neural networks (CNNs) mainly used for sequence data to extract local features in sequences through convolutional layers or graph convolutional networks (GCNs)mainly used for structural data to capture topological information in protein structures through graph convolutional layers\cite{RN49}. However, existing computational methods have significant shortcomings in capturing the hierarchical geometric relationships among proteins and the multiscale dynamic patterns of their interactions\cite{xu2023new}. Among them, graph convolutional networks usually rely on graph neural networks in Euclidean space, where the volume of the ball grows polynomially with the radius, and cannot effectively express the exponentially growing scale-free properties and hierarchical structure in the network\cite{RN27}. Due to the above reasons, graph node vectors embedded in Euclidean space will suffer from large distortions in representing the network structure, especially for the networks with hierarchical or tree-like structures For the above reasons, the node vectors embedded into the Euclidean space have large distortion in representing the network structure, especially complex networks with hierarchical or tree-like structures, and the traditional GNN is inefficient in message passing due to the phenomenon of “over-squashing”\cite{RN48} when dealing with large-scale graphs, which makes it difficult to capture the relationship between remote nodes.
To solve some of the problems of graph neural networks in Euclidean space, we propose to embed the graph into a hyperbolic space, where the volume of the ball grows exponentially with the radius, which can better express the exponential growth characteristics of the nodes in the network and the scale-free distribution, and the hyperbolic space can naturally embed the graph with a hierarchical structure, preserving the power distribution, strong clustering, and small-world characteristics of the network, to obtain a more accurate node vector representations\cite{RN49}, and the geometric flow through the hyperbolic space allows for graph learning and evolution in continuous time, and improves computational efficiency using efficient numerical solvers. Meanwhile, the introduction of random wandering matrix-based GWT can analyze the graph signals at different scales, capturing both local and global information on node features. This is particularly important for dealing with biological networks with complex hierarchical structures, as it can reveal the interaction patterns at different levels\cite{RN51}. The “over-squashing” problem is also alleviated by random wandering diffusion\cite{RN50}, and different levels of information are hidden at different scales. For example, in biological networks, a small scale may reflect the close relationship between local nodes, while a large scale may reflect the overall modular structure or the connection between global functional units. This multi-scale analysis helps to comprehensively capture feature information in graph data, which is difficult to do with single-scale methods.
\section{Materials and Methods}
\subsection{Dataset}
We used the publicly available heterogeneous network dataset proposed by Luo as the experimental dataset\cite{RN64}. This dataset was extracted from the HPRD database (Release 9) for protein nodes, and protein-protein interactions were downloaded from the HPRD database (Release 9). In addition, we excluded those isolated nodes; in other words, we only considered nodes with at least one edge in the network. In terms of dividing the dataset 85\% of the training set, 5\% of the validation set, and 10\% of the test set were chosen.

\subsection{HyboWaveNet Model}

\begin{figure}[h!tbp]
    \centering
    \includegraphics[width=\textwidth]{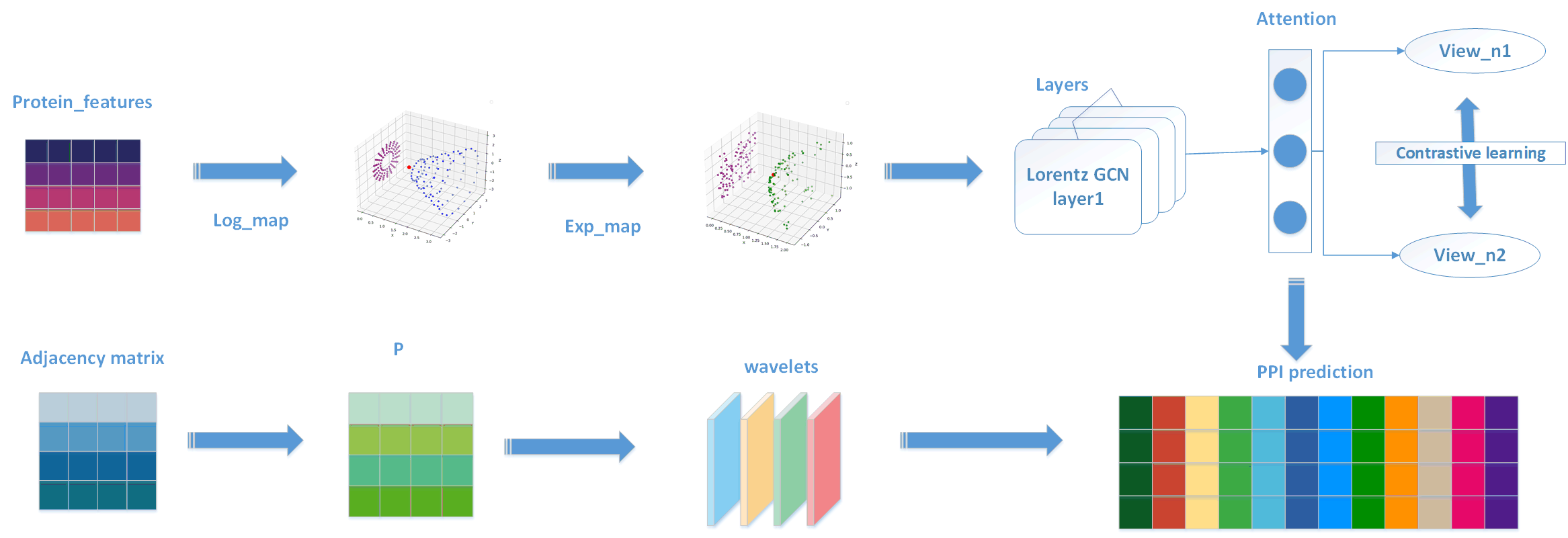}
    \caption{HyboWaveNet workflow. (a) Project the feature matrix into the tangent space, and then project from the tangent space into the hyperbolic space to generate the feature matrix under the hyperbolic space, and input it into the multilayer GCN to do comparison learning by using the attention mechanism to aggregate the features extracted from each layer of GCN. (b) Calculate the random wandering matrix P by the adjacency matrix, then derive the wavelet coefficients from P and the scale list, and finally multiply them with the feature matrix aggregated by the attention to get the final features to obtain the PPI prediction.}
    \label{fig:model_all}
\end{figure}

\subsubsection{ Lorentz space based graph neural network encoder:}
Graph neural networks are commonly applied to extract features from graph-structured data (15), and he can aggregate features from neighboring nodes. Through the aggregation of information from neighbor nodes, graph neural networks can effectively extract local features, however, traditional graph neural networks operate in Euclidean space, which makes it difficult to effectively capture data with a hierarchical or tree structure, but hyperbolic space has the property of exponential expansion, which can more naturally represent hierarchical or tree-structured data, and in hyperbolic space, the distances between nodes can better reflect their semantic relationships, especially in hierarchical structures. Thus, a graph neural network encoder based on hyperbolic geometry is introduced. Hyperbolic space can naturally capture hierarchical structure, while it can effectively aggregate neighbor information, and the combination of the two can better model graph data with hierarchical structure.
The hyperbolic space\cite{chen2021fully}of the Lorentz model is defined as:
\begin{equation}\label{eq:inner_product}
L^n = \{x \in R^{n+1} \mid \langle x, x \rangle_{\tau} = -1, x_0 > 0\}
\end{equation}
where $\langle \mathbf{x},\mathbf{y} \rangle_{\tau}$ is the Lorentz inner product
\begin{equation}\label{eq:inner_product}
\langle x, y \rangle_{\tau} = x_{0}y_{0} + \sum_{i=1}^{n} x_{i}y_{i}
\end{equation}
The core of HyboNet is LorentzGraphconvolution that performs neighbor aggregation and feature transformation in hyperbolic space.
First for node v, its neighbor aggregation can be expressed as
\begin{equation}\label{eq:inner_product}
\mathbf{h}_v^{(l)} = Aggregate(\{ \mathbf{h}_u^{(l-1)} \mid u \in N(v) \})
\end{equation}
where $\mathbf{h}_{v}^{(l)}$ is the embedding of node $v$ in layer $l$, 
$\mathcal{N}(v)$ is the set of neighbors of node $v$, and Aggregate is the aggregation function.
After aggregation in hyperbolic space, the linear transformation is realized by the Lorentz transformation.
\begin{equation}
\mathbf{h}_{v}^{(l)} = \exp_{o}^{c}\left(W\log_{o}^{c}(\mathbf{h}_{v}^{(l-1)}) + b\right)
\end{equation}
where $\exp_{\mathbf{o}}^{c}$ and $\log_{\mathbf{o}}^{c}(\mathbf{x})$are the exponential and logarithmic mappings in the Lorentz model, and W and b are the learnable weight matrix and bias vector. Finally the features are obtained by hyperbolic activation function
\begin{equation}
\sigma = \exp_{o}^{c}\left(\sigma_{\text{Euclidean}}\log_{o}^{c}(x)\right)
\end{equation}
where $\sigma_{Euclidean}$ is the activation function in Euclidean space and $\exp_{\mathbf{o}}^{c}$ and $\log_{\mathbf{o}}^{c}(\mathbf{x})$ are the exponential and logarithmic mappings in the Lorentz model.
HyboNet is a graph neural network encoder based on Lorentz's hyperbolic space, and its forward propagation process can be formalized as the following cascade operation:
\begin{equation}
\mathbf{H}^{(l)} = \sigma_{L}\left(Aggregate_{L}\left(\mathbf{H}^{(l-1)}, A; W^{(l)}, c^{(l)}\right)\right)
\end{equation}
where \( H^{(l)} \in {L}^{d^{(l)}} \) is the node embedding matrix of layer \( l \) belonging to the Lorentz hyperbolic space with curvature \( c^{(l)} \), \( A \in \{0,1\}^{N \times N} \) is the adjacency matrix of the graph, \( W^{(l)} \) is the learnable weight matrix of layer \( l \), \( \sigma_{L} \) is the nonlinear activation function in the hyperbolic space, and \( \text{Aggregate}_{L} \) is the neighbor aggregation operation in the hyperbolic space.
\subsubsection{Multi-scale Spatial Wavelet Transform via Random Walk Diffusion}
The traditional method is based on the graph Laplacian matrix $\mathbf{L}= \mathbf{D}-\mathbf{A}$, whose spectral decomposition is $\mathbf{L}=\mathbf{U}\mathbf{\Lambda}\mathbf{U}^{T}$. Instead, multiscale neighborhood relationships are explicitly modeled by random wandering null-space diffusion to avoid the computational bottleneck of feature decomposition. The number of diffusion steps $s$ directly corresponds to the spatial range from local ($s=1$) to global ($s=K$), which is easier to align with biological hierarchies (e.g., residue-structural domain-protein).

The first step is the construction of the random wandering matrix. Given an undirected graph $\mathbf{g}=(\mathbf{v},\boldsymbol{\varepsilon})$ and an adjacency matrix $\mathbf{A}\in\{0,1\}^{N\times N}$, construct the adjacency matrix with self-loop:
\begin{equation}
\tilde{\mathbf{A}} = \mathbf{A} + \mathbf{I}_N
\end{equation}
The degree matrix $\mathbf{D}\in{R}^{N\times N}$ is defined as a diagonal matrix whose elements satisfy:
\begin{equation}
D_{ii} = \sum_{j=1}^{N}\tilde{A}_{ij}
\end{equation}
The random wandering matrix $\mathbf{P}\in{R}^{N\times N}$ is defined as:
\begin{equation}
\mathbf{P} = \mathbf{D}^{-1} \tilde{\mathbf{A}}
\end{equation}
Afterwards, generate a multiscale diffusion operator based on the randomized wandering matrix and the list of scales.
Given a set of scale parameters $\mathcal{S} = \{s_{1}, s_{2}, \dots, s_{K}\}$, define the scale-dependent diffusion operator $\mathbf{T}_{s}\in R^{N\times N}$ as a power of the random walk matrix:
\begin{equation}
\mathbf{T}_{s} = P^{s}
\end{equation}
where $s\in\mathcal{S}$ controls the number of diffusion steps and larger $s$ captures more global neighborhood information.
Finally, the generated wavelet coefficients are used with the feature matrix for graph wavelet feature extraction.
Given the node feature matrix $\mathbf{X}\in R^{N\times d}$, the multiscale graph wavelet feature $\mathbf{Z}\in R^{N\times (K\cdot d)}$ is defined as:
\begin{equation}
\mathbf{Z} = \|_{s\in \mathcal{S}}\mathbf{T}_{s}\mathbf{X}
\end{equation}
where $\|$ denotes splicing by columns and the final features contain the diffusion signals at all scales.
Overall, given the adjacency matrix $\mathbf{A} \in \{0,1\}^{N \times N}$ of the graph $\mathbf{g} = (\mathbf{v}, \boldsymbol{\varepsilon})$, the node feature matrix $\mathbf{X} \in R^{N \times d}$, and the set of scale parameters $\mathcal{S} = \{s_1, s_2, \cdots, s_K\}$, the multiscale graph wavelet transform is defined as
\begin{equation}
\mathbf{Z} = \|_{s \in \mathcal{S}} \left( \mathbf{D}^{-1} (\mathbf{A} + \mathbf{I}_N) \right)^s \mathbf{X}
\end{equation}
where $\mathbf{D} \in {R}^{N \times N}$ is the degree matrix with self-loop that satisfies $D_{ii} = \sum_{j=1}^{N} (\mathbf{A} + \mathbf{I}_N)_{ij}$; $\mathbf{T}_s = \left( \mathbf{D}^{-1} (\mathbf{A} + \mathbf{I}_N) \right)^s$ is the diffusion operator of scale $s$, denoting the $s$-step random wandering probability matrix, and $\| \cdot \|$ denotes the collocation operation by columns, with the final output $\mathbf{Z} \in R^{N \times (K \cdot d)}$ containing features of all scales.
\subsubsection{Contrastive Learning Module}
In the contrast learning module, we take the features extracted by HyboNet after randomly dropping (dropout) 20\% to form two features in different viewpoints as a pair of positive sample pairs, and other nodes are defined as negative sample pairs. The distance is calculated by cosine similarity.

Let $\mathbf{z}_1^{(i)} \in {R}^{d}$ and $\mathbf{z}_2^{(i)} \in R^{d}$ denote the embedding vectors of node $i$ under two different data augmentation (or dropout) perspectives, respectively. $\tau$ is a temperature hyperparameter (with a default value of 0.1), which is used to regulate the distribution of similarity. The embedding vectors are first L2 normalized:

\begin{equation}
\mathbf{z}_1^{(i)} = \frac{\mathbf{z}_1^{(i)}}{\|\mathbf{z}_1^{(i)}\|_2}, \quad
\mathbf{z}_2^{(i)} = \frac{\mathbf{z}_2^{(i)}}{\|\mathbf{z}_2^{(i)}\|_2}
\end{equation}
where $\|\cdot\|_2$ denotes the L2 norm.
Compute the cosine similarity matrix between two viewpoint embeddings:
\begin{equation}
\text{sim}\left(\mathbf{z}_1^{(i)}, \mathbf{z}_2^{(j)}\right) = \frac{\mathbf{z}_1^{(i)} \cdot \mathbf{z}_2^{(j)}}{\tau}
\end{equation}
where denotes dot product operation.
Positive sample pairs: two viewpoint embeddings ($\mathbf{z}_1^{(i)}, \mathbf{z}_2^{(i)}$) of the same node $i$,
Negative sample pairs:embeddings$(\mathbf{z}_1^{(i)},\mathbf{z}_2^{(j))}$of different nodes $i\neq j$.
\begin{equation}
L_{\text{contrastive}} = -\frac{1}{N} \sum_{i=1}^{N} \log \frac{\exp\left( \text{sim}(\mathbf{z}_{1}^{(i)}, \mathbf{z}_{2}^{(i)}) \right)}{\sum_{j=1}^{N} \exp\left( \text{sim}(\mathbf{z}_{1}^{(i)}, \mathbf{z}_{2}^{(j)}) \right)}
\end{equation}
where the numerator is the similarity of the positive sample pairs, the denominator is the sum of the similarity of the positive sample pairs with all the negative sample pairs, and $N$ is the number of nodes.
\subsection{Protein-Protein Interaction Prediction}
Here the feature representation obtained by the encoder is used to calculate the squared distance in Lorentz hyperbolic space between two nodes as the node interaction score:
\begin{equation}
\text{score}(v_i, v_j) = -\text{sqdist}_{\mathcal{L}}(\mathbf{h}_{v_i}, \mathbf{h}_{v_j}; c)
\tag{16}
\end{equation}
where $\mathbf{h}_{v_i}, \mathbf{h}_{v_j} \in {L}^{n}$ are the embeddings of nodes $v_i$ and $v_j$ in Lorentz hyperbolic space,$\textbf{sqdist}_{\mathcal{L}}(\cdot)$ is the squared distance function in Lorentz hyperbolic space, and $c > 0$ is the curvature parameter of Lorentz space.
\section{Experiments and Results}
\subsection{Benchmark Comparison}
\subsubsection{Experimental Setup}
PPI can be understood as a binary classification task, so the area under the ROC curve (roc) and the area under the precision-recall curve (ap) are used as evaluation metrics. The positive samples used for PPI prediction are chosen from our known PPIs and thenegative samples are chosen from unknown PPIs.
We evaluated HyboWaveNet and the baseline method on the same dataset with the following settings. In HyboWaveNet, we set the number of scales of the multiscale wavelet transform to 4, 1, 2, 3, and 4. Comparison of the different views in the learning was generated with a random discard of 20\%. The learning rate was set to 1e-3 and the temperature parameter was set to 0.2. 2000 runs were performed with the Adam optimizer with early stopping mechanism.
\subsubsection{Results and Comparative Analysis}
Comparison with benchmarks
We will compare with the following 4 state-of-the-art benchmark methods (shown in Table \ref{tab:results}), for fairness we choose the same dataset and use the area under the ROC curve (roc) and the area under the precision-recall curve (ap) as the evaluation metrics, and it can be seen that our model outperforms all the other 5 models:
AGAT\_PPIS\cite{RN60}is a protein-protein interaction site predictor based on initial residuals and homogeneity mapping of augmented graph attention networks.
The GACT\_PPIS\cite{RN61}algorithm is designed to predict protein-protein interaction sites using combined protein sequence and structure information as input.
Struct2Graph\cite{baranwal2020deep}is a structure-based prediction method. It converts 3D protein structures into graphs and applies graph-based learning to infer interactions.
Topsy\_Turvy\cite{RN59}is a sequence-based approach that uses amino acid sequences and evolutionary information to predict interactions. And uses migration learning techniques to enhance the prediction performance based on sequence data.
Fully\_HNN is a fully hyperbolic framework based on the Lorentz model for building hyperbolic networks by employing Lorentz transformations (including boost and rotation) to formalize the basic operations of neural networks.
\begin{table}[htbp]
\centering
\caption{Comparison of performance with benchmark models}
\label{tab:results}
\renewcommand{\arraystretch}{1.0} 
\begin{tabular}{l@{\hspace{1.5cm}}c@{\hspace{1.5cm}}c}
\toprule
Model & AUC & AUPR \\
\midrule
AGAT\_PPIS & 0.866 $\pm$ 0.004 & 0.582 $\pm$ 0.002 \\
GACT\_PPIS & 0.878 $\pm$ 0.005 & 0.594 $\pm$ 0.006 \\
Topsy\_Turvy & 0.868 $\pm$ 0.003 & 0.867 $\pm$ 0.004 \\
Struct2Graph & 0.892 $\pm$ 0.001 & 0.873 $\pm$ 0.001 \\
Fully\_HNN & 0.909 $\pm$ 0.002 & 0.926 $\pm$ 0.003 \\
HybowaveNet & 0.922 $\pm$ 0.005 & 0.938 $\pm$ 0.005 \\
\bottomrule
\end{tabular}
\end{table}
\subsection{Ablation experiments}
\begin{figure}[h!tbp]
    \centering
    \includegraphics[width=\textwidth]{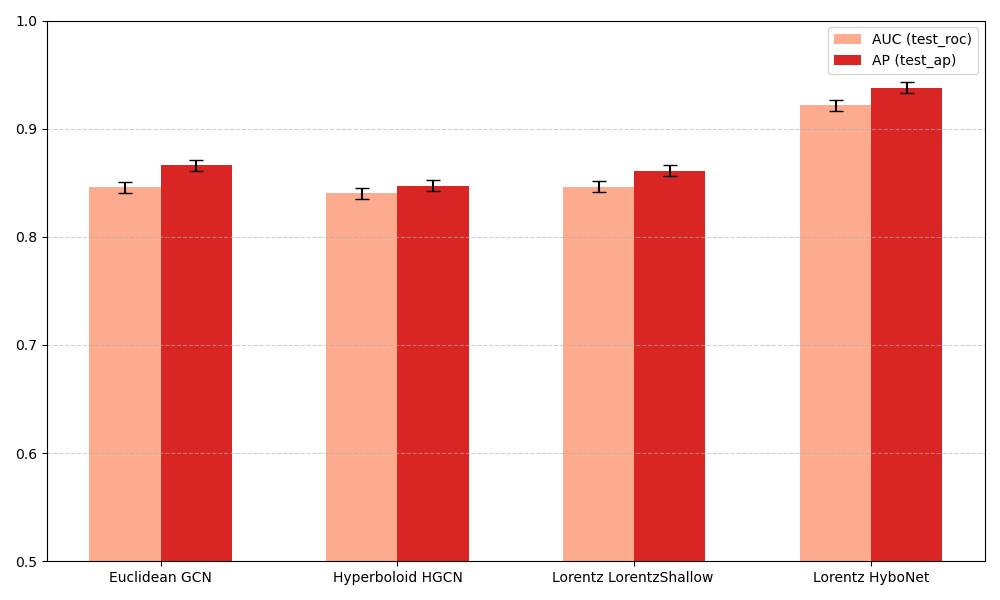}
    \caption{Values of the two assessment metrics, AUC and AP, with different encoders}
    \label{fig:change_encoder}
\end{figure}

\begin{figure}[h!btp]
    \centering
    \includegraphics[width=0.9\textwidth]{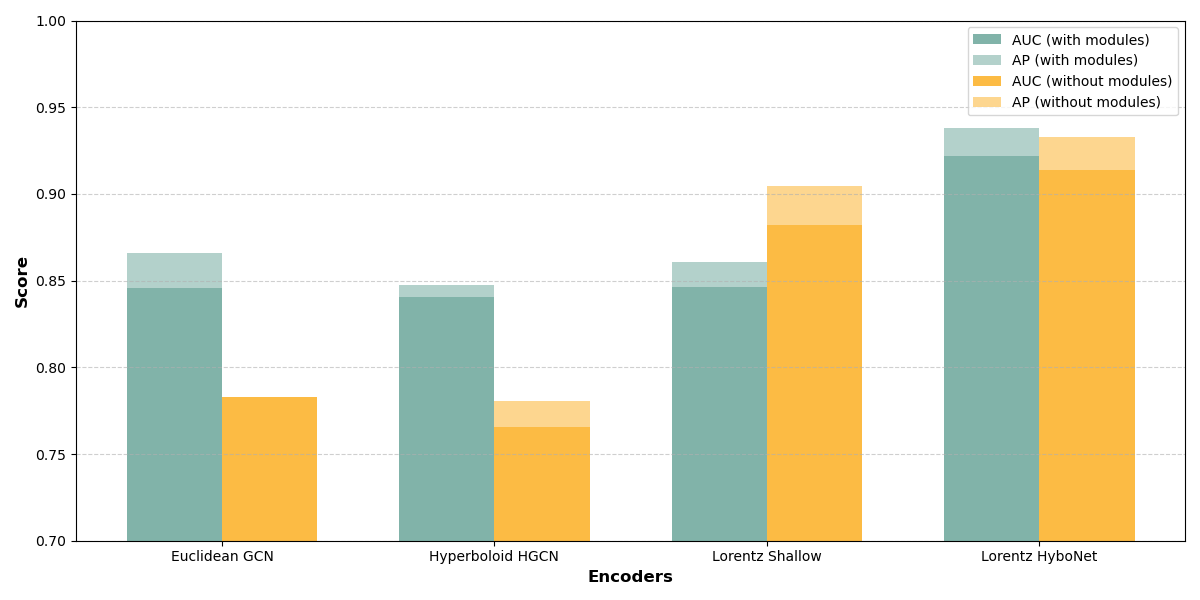}
    \caption{Values of AUC and AP for two evaluations with different encoders with and without multiscale wavelet transform and comparison learning module}
    \label{fig:gwt}
\end{figure}

For the ablation of the graph neural network encoder based on Lorentz space, with scale [1, 2, 3, 4] and temperature parameter 0.2, we changed the encoder to GCN under Euclidean, HGCN under Hyperboloid, and LorentzShallow under Lorentz for the experiments as shown in Fig. \ref{fig:change_encoder} , and after the The values of AUC and AP are decreased after changing the encoder.
For the ablation experiments based on random wandering diffusion with multi-scale nullspace wavelet transform and contrast learning, the encoders were added to the multi-scale graph wavelet transform and contrast learning modules in GCN under Euclidean, HGCN under Hyperboloid, LorentzShallow under Lorentz, and HyboNet under Lorentz, respectively and deletion. Conducting the experiments as shown in Fig. \ref{fig:gwt} it can be observed that except for the LorentzShallow encoder under Lorentz the values of both evaluation metrics decreased after the absence of the graph wavelet transform and contrast learning, in particular, the AP decreased by 0.0861 in the GCN under Euclidean, and the AUC in the HGCN under Hyperboloid decreased by 0.0747.
Based on the results of the above two experiments, we conclude that both the HyboNet under Lorentz and the multiscale wavelet transform under random wandering diffusion and the contrast learning module that we use can effectively improve the prediction performance.
\subsection{Hyperparameter Sensitivity Analysis}
\begin{figure}[H]
    \centering
    \includegraphics[width=\textwidth]{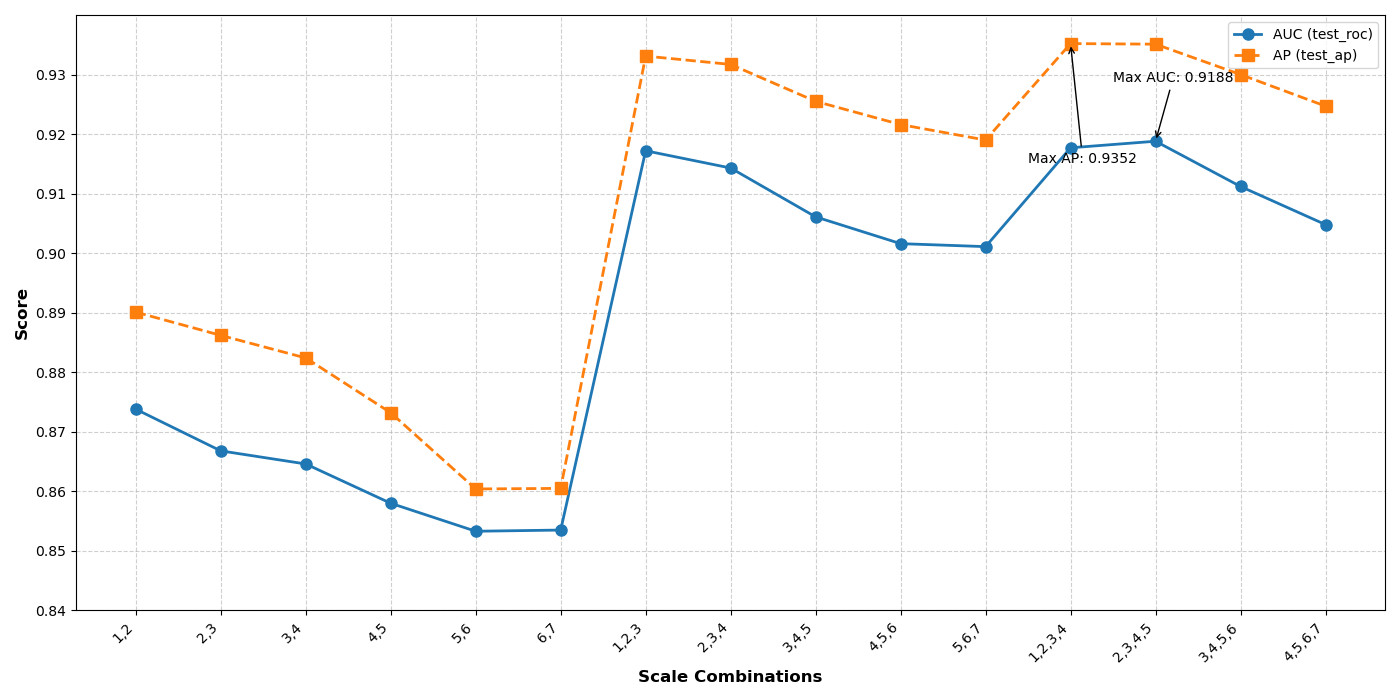}
    \caption{Values of AUC and AP under different scale lists}
    \label{fig:scale}
\end{figure}

We have performed sensitivity analysis of hyperparameters of HyboWaveNet, to find the best case we first consider the size of the number of scales and the value of each scale in the graph wavelet transform. Consider the number of scales from 2, 3, 4 and the scale value is not greater than 7. As shown in Fig.\ref{fig:scale} it is found that the number of scales from 2 to 3 the performance is greatly enhanced so we consider that the number of scales 3 and 4 can learn a wider and deeper structure. The AUC and AP tend to stabilize at a scale number of 3 and a scale list of (1, 2, 3, 4), and finally we believe that the model can better learn hierarchical topological relationships among biomolecules under this scale criterion.
\begin{figure}[h!tp]
    \centering
    \includegraphics[width=0.9\textwidth]{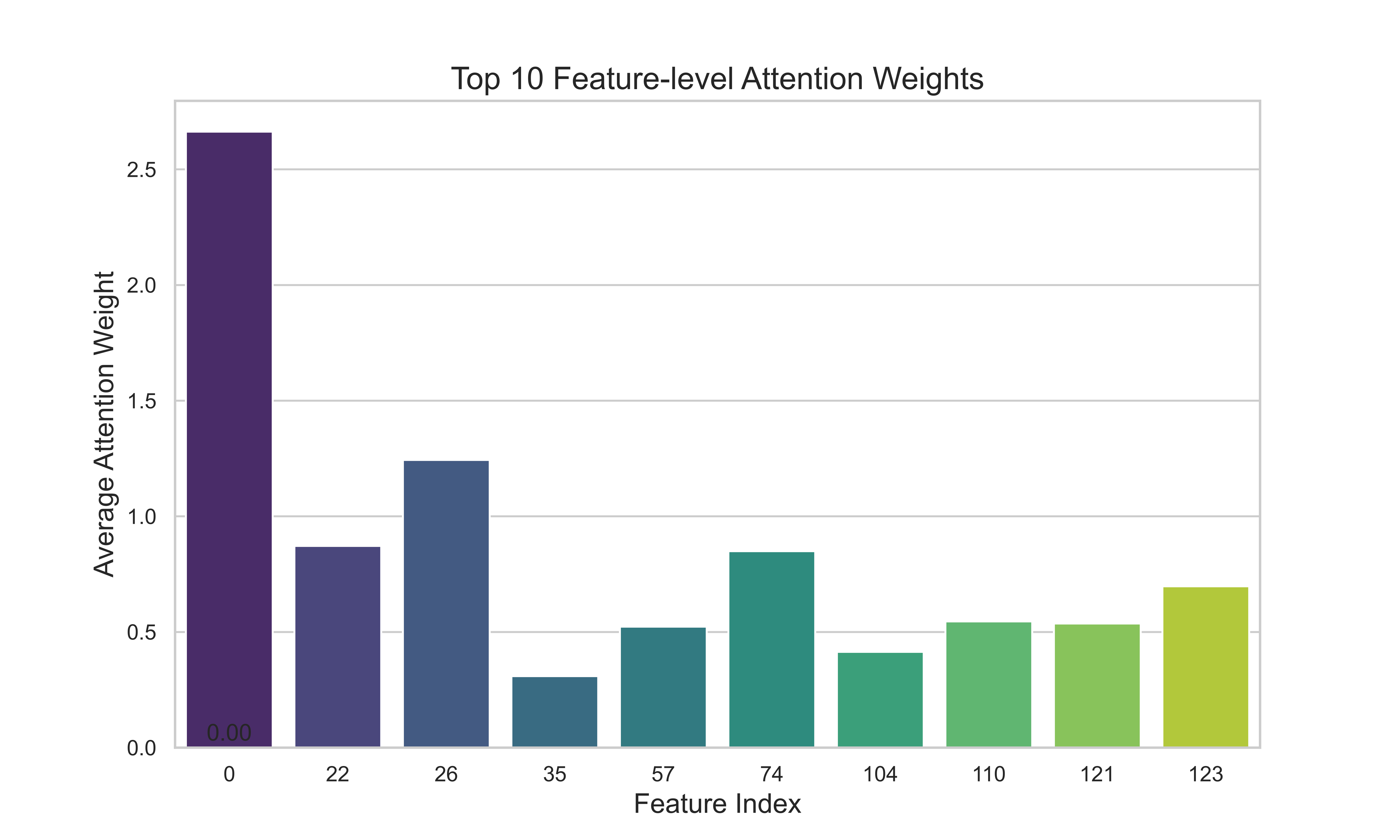}
    \caption{The 10 features in the initialized node features that contribute the most to PPI prediction}
    \label{fig:attention}
\end{figure}
\section{Conclusion and Future Work}
In this work,we find out the 10 features in the initialized 128-dimensional feature vector that contribute the most to the PPI prediction through the allocation of attention weights, as shown in Fig.\ref{fig:attention} . we overcame the limitations of graph-based models by combining hyperbolic geometry and multiscale signal processing techniques to enhance the accuracy of PPI prediction, providing a novel and effective approach for PPI prediction. It provides a new perspective for research in the field of bioinformatics. Meanwhile, it is found that some features in the initialized 128-dimensional feature vector contribute a lot to the PPI prediction. 

\clearpage
\footnote{This work was supported by the Science and Technology Project of Jiangxi Provincial Department of Education, GJJ2201043.}

\end{document}